\documentclass[12pt]{article}

\usepackage{times}
\usepackage{epsfig}
\usepackage{graphicx}
\usepackage{amsmath}
\usepackage{amssymb}
\usepackage{latexsym,amsfonts, amsbsy}
\usepackage{algorithmic}
\usepackage{setspace}

\begin{document}

\title{A Invertible Dimension Reduction of Curves on a Manifold}

\author{Sheng Yi\\
North Carolina State University\\
 \and
Hamid Krim\\
North Carolina State University\\
\and Larry K. Norris\\
North Carolina State University}

\maketitle

\begin{abstract}\label{sec.abs}
In this paper, we propose a novel lower dimensional representation of a shape
sequence. The proposed dimension reduction is invertible and computationally more efficient in comparison to other related works. Theoretically, the differential geometry tools such as moving frame and parallel transportation are
successfully adapted into the dimension reduction problem of high dimensional curves. Intuitively, instead of searching for a global flat subspace for curve embedding, we deployed a sequence of local flat subspaces adaptive to the geometry of both of the curve and the manifold it lies on. In practice, the experimental results of the dimension reduction and reconstruction algorithms well illustrate the advantages of the proposed theoretical innovation.
\end{abstract}

\section{Introduction}\label{sec.intro}

For many pattern recognition problems, the feature space is
nonlinear and of high dimensionality and nonlinearity, for example,
face recognition, image classification and shape recognition. Many
dimension reduction techniques \cite{F02} are therefore naturally
introduced to ease the modeling, computation and visualization of
the feature space. Dimension reduction has been widely used
with different vision techniques such as contour tracking
\cite{B95}, face recognition \cite{CHT03} and object recognition
\cite{AK10}.

One of the key points of dimension reduction is how to represent
the geometry of the original space in a lower dimensional space
\cite{TSL00,RS00}. Most of the previous dimension reduction research
\cite{F02} focuses on embedding the points in the original space
into a lower dimensional space. Beyond the point-wise dimension
reduction, the dimension reduction for a sequence of points or an evolving curve
in the original feature space is more important in applications
where features of interest are time varying such as, face expression
\cite{CHT03}, events indexing in video sequence as a shape process
\cite{HNB04} and human activity as a shape sequence \cite{B95}. To the best of our knowledge, only a few works
\cite{B95,AK10} address the problem of how to represent a curve in the original feature space in a
lower dimensional space.

The dimension reduction of high dimensional curves may be
categorized into two classes. One is to reduce the whole curve space
to a lower dimension as in \cite{B95,BCZ93,CB92}. The other one is
to reduce a particular the curve to a
corresponding lower dimensional curve as in \cite{AK10}. The
drawback of the first one is the Euclidean assumption by using an
approximation with a spline function. Moreover, as in \cite{B95} by
directly applying PCA to the spline coefficients, the global
dimension reduction does not consider the intrinsic dimensionality
of a particular high dimensional curve. A curve sequence like walking, for
example, is much simpler than a dancing sequence. Thus the dynamics
of walking might be represented in a much lower dimensional space in
comparison to other complex activities. The second category of
dimension reduction as in \cite{AK10}, considered the geometry of a
particular curve in the original space, which, due to the chosen
implementation of the Whitney embedding theorem, is computationally
heavy on account of the iterative subspace search. At the same time,
there exists no invertible mapping from the embedding space to the
original one, thus limiting its extension to a generative process.

Motivated by the importance of a lower dimensional representation of
a manifold-valued curve and the limitations of the current
techniques, in this paper a novel dimension reduction of a shape
manifold valued curve is proposed. In contrast with  previous curve
dimension reduction techniques \cite{B95,AK10}, the proposed method
has the following advantages:
\begin{enumerate}
\item the dimension reduction is adaptive to the Riemannian geometry of a shape space.
\item the computation is linear in the length of the curve times the dimension of the shape space.
\item with a proper dimensionality, there exists a reconstruction
as a subspace parallel transport on a manifold.
\end{enumerate}

To introduce the idea of the proposed dimension reduction framework
for the manifold valued curves, this paper follows the
moving frame formulation in \cite{EH59,KN96}. According to \cite{KN96,EH59},
every curve on a manifold may be represented by a
differential equation,
\begin{equation}
X_{t} = X_{0} + \int_{0}^{t}
\sum_{i=1}^{dim(M)}V_{i}(X_{t})dZ_{i}(t)
\end{equation}
where, $V(X_{t})$ is a vector field on a manifold, and $Z_{t}$ is
a curve in a Euclidean space. Under such representation, to
introduce a lower dimensional representation of $X_{t}$, a method is
proposed to find the optimal vector field $V(X_{t})$ such that:
\begin{enumerate}
  \item The sequence of $V(X_{t})$ is path dependent and uniquely
  determined by $X_{t}$ and initial condition $V(X_{0})$ by a efficient computation.
  \item The resulting $dZ_{t}$ from $X_{t}$ and $V(X_{t})$ have a distribution that lies in a lower
dimensional subspace.
\end{enumerate}

The first requirement on $V(X_{t})$ means that $X_{t}$ can be
represented by only the initial condition $V(X_{0}),X_{0}$ and the
embedded curve $Z_{t}$. According to \cite{EH59} such
property yields the 1-1 correspondence between $X_{t}$ and $Z_{t}$. The second requirement is more
intuitive that among all the $V(X_{t})$ satisfied the first
requirement, the optimal one is selected such that the variance of
tangents is best represented in a subspace.

The first requirement is satisfied by select $V(X_{t})$ that is a horizontal vector field \cite{KN96}
defined by the Levi-Civita connection as in Section \ref{sec.MFG} with a
$L^{2}$ metric induced to the shape manifold $M$ from the ambient
space. According to \cite{EH59,KN96}, under a defined connection, for $X_{t} \in M$ and initial condition
$V(X_{0})$, there exists a unique selection of moving frame $V(X_{t})$. Thus given $V(X_{0})$ the resulting curve in Euclidean space have one-one correspondence to $X_{t}$. The reason of
using the Levi-Civita connection have two folds. The first one is
that it is a non-flat connection. In contrast to the other category:
flat connection, it is path dependent. Such property is important
for dimension reduction because it provides more adaptivity to the
geometry of the path on the manifold. The second reason is that in
comparison with the general nonflat connections defined by a PDE
on manifold, such particular Levi-Civita connection in Section \ref{sec.MFG} is consistent with the metric of the shape manifold in \cite{KS04} that this paper is based on, and is computationally more efficient for a known ambient space and normal space of
the manifold, as is the case of our shape manifold.

The second requirement is achieved by the subspace representation
resulting from maximizing the variance of the corresponding curve development $dZ(t)$ with different choices of $V(X_{0})$. In
theory, without knowing the initial condition $V(X_{0})$ the
Levi-Civita connection only determines the vector field $V(X_{t})$
up to a group action within the fiber of the Frame bundle (the total space of frames of the tangent space of a manifold) \cite{KN96}. Among
different choices for $V(X_{t})$ along the fiber in the frame bundle, a optimal frame is selected
such that the previously discussed second requirement is satisfied
to make the resulted $Z{t}$ better representable in a lower
dimensional space. A $\tilde{V}(X_{0})=
\{x_{0},e_{1},e_{2},\cdots,e_{dim(M)}\}$ is selected by carried out PCA
of all the parallel transported tangents along $X_{t}$.let $\{e_{1},e_{2},\cdots,e_{L}\}$ are the eigenvectors
corresponding to the $L$ largest Eigenvalues in the PCA calculation. Then A subspace spanned by the subset $\{e_{1},e_{2},\cdots,e_{L}\}$ is constructed to better represent the variance of all the tangents along $X_{t}$ in a lower dimensional
space.

To summarize, a curve in the shape manifold $M$ is represented as
\begin{equation}
X_{t} = X_{0} + \int_{0}^{t} \sum_{i=1}^{L}
\tilde{V}_{i}(X_{s})d\tilde{W}_{i}(s)
\end{equation}
where $\tilde{V}_{i}(X_{t})$ can be represented by $L$ curves in
$R^{3}$ and $\tilde{Z}_{i}(t)$ is represented by a curve in $R^{L}$.
Our proposed dimensional reduction achieves the following points,

\begin{enumerate}
  \item a lower dimensional representation of curves on a
  manifold.
  \item The moving frame representation can be generated to a stochastic representation of random process on a manifold. Thus the propose method can be
  well adapted to a generative modeling in a lower dimensional space.
  \item the computation is linear in the dimension of $X_{t}$.
  \item there exists a reconstruction from the lower dimensional
  representation to the original curve on the shape manifold.
\end{enumerate}

The rest of the paper is organized as following. Section \ref{sec.pre} provide necessary background knowledge of shape manifolds and the geometric calculation on it. In Section \ref{sec.dim_red} and \ref{sec.dim_rec}, the dimension reduction framework is introduced,
which includes the curve embedding and the curve reconstruction.

\section{Preliminary}\label{sec.pre}
In this section, we first introduce in Subsection \ref{sec.shape_manifold} the shape manifold that the dimension reduction is based on. Then in Subsection \ref{sec.MFG} a fundamental moving frame representation of curves on manifold is briefly introduced.

\subsection{Shape Manifold}\label{sec.shape_manifold}
According to \cite{KS04}, a planar shape is a simple and closed
curve in $\mathbb{R}^{2}$,
\begin{equation}\label{eq.CurFunc}
\alpha(s): I \rightarrow \mathbb{R}^{2},
\end{equation}
where an arc-length parameterization is adopted.  A shape is
represented by a \emph{ direction index function} $\theta(t)$.  With
such a parameterization,  $\theta(s)$  may be associated to the
shape by
\begin{equation}\label{eq.TanFunc}
\frac{\partial \alpha}{\partial s} = e^{j \theta(s)}.
\end{equation}

The ambient space of the manifold of $\theta$ is an  affine space
based on $\mathbb{L}^{2}$. Thus
\begin{equation}\label{eq.AfSp}
\theta \in A(\mathbb{L}^{2}).
\end{equation}

The  restriction of a shape is that it must be a closed and simple
curve, and invariant over rigid Euclidean transformations. The shape
manifold $M$ is defined by a level function $\phi$ as
\begin{equation}\label{eq.ManFunc}
\phi(\theta) = \left(\int_{0}^{2\pi} \theta ds, \int_{0}^{2\pi}
\cos(\theta) ds, \int_{0}^{2\pi} \sin(\theta) ds\right).
\end{equation}
\begin{equation}\label{eq.ManDef}
M = \phi^{-1}(\pi,0,0).
\end{equation}

One of the most important properties of $M$ is that the tangent
space $TM$ is well defined.  Such a property not only simplifies the
analysis, but also makes the incremental computation possible,
\begin{equation}\label{eq.ManDef}
T_{\theta}M = \{f\in \mathbb{L}^{2}| f~\bot
~span\{1,\cos(\theta),\sin(\theta)\}\}.
\end{equation}

In addition, an iterative projection is proposed in \cite{KS04} to
project the point in ambient space back to the shape manifold $M$.
The idea is that each time $\theta$ is updated as $\theta+d\theta$,
where $d\theta$ is orthogonal to the level set
$\phi^{-1}(\phi(\theta))$. The $d\theta$ is calculated as
$d\phi^{-1}((\pi,0,0)-\phi(\theta))$. For the detailed form of the
Jacobian of $d\phi$, one could refer to \cite{KS04}.

The problem of this manifold for our  stochastic modeling is the
infinite dimension. The mapping of random process in \cite{EH59}
from a manifold to a flat space is only defined on a finite
dimensional manifold. Therefore, a Fourier approximation of the shape manifold we discussed above is
developed, such that the dimension is reduced to a finite number.

\subsection{Moving Frame Geometry}\label{sec.MFG}
Let $X_{t}$ be a curve on a manifold $M$, then following the moving frame representation \cite{BC64}, the tangent $dX_{t}$ of manifold valued curve $X_{t}$
may be written as:
\begin{equation}
dX_{t} =  \sum_{i}V_{i}(X_{t})dZ_{t},
\end{equation}
where $\{V_{i}(X_{t})\}_{i=1, 2, \cdots, dim(M)}$ is a frame of the tangent space at $X_{t}$, which is denoted as $T_{X_{t}}M$. Consequently $dZ_{t}$ may be understood as a linear coefficient of $dX_{t}$ under the representation of $V(X_{t})$. Such moving frame representation is widely used in geometry studies of curves. For example in the Frenet theorem \cite{BC64} the geometry of curve is uniquely identified by the curvature of $V(X_{t})$ up to a rigid Euclidean transformation. Recently in computer graphics, a rotation minimizing moving frame \cite{WJ97} are developed to avoid the singularity of original moving frame used in Frenet theorem.

In this paper, the moving frame representation is utilized for the similar purpose of representing a curve on a manifold. In contrast to the previous  designs of moving frame, in this paper we elaborate to propose a adaptive moving frame to represent a high dimensional curve in a lower dimensional space. Vector field $V(X(t))$ are developed as a sequence of parallel frames along $X_{t}$. The parallelism is defined under a Levi-Civita connection. The advantages of such innovation are briefly introduced in the following:
\begin{enumerate}
\item Once $V(X_{t})$ are parallel according to the curve development theory in \cite{EH59}, $X_{t}$ can be represented uniquely as $(X_{0}, V_{0},Z_{t})$. Such property yields the invertibility of the proposed dimension reduction of $X_{t}$. In other words, by the representative curve $Z_{t}$ in a lower dimensional space, $X_{t}$ can be reconstructed when the initial conditions $(X_{0},V_{0})$ are available.
\item Under Levi-Civita connection, the parallel frames are calculated according to metric of the manifold $M$. Such property implies that the shape of $Z_{t}$ will reflect well the shape of $X_{t}$ on $M$. For example, if $X_{t}$ is a geodesic curve, then in the proposed method $Z_{t}$ will be a straight line in Euclidean space
\item The particular shape manifold \cite{KS04} we adopted in this paper have a known tangent and normal space in the ambient space which greatly simplified the calculation of the moving frames under Levi-Civita connection.
\end{enumerate}

The Levi-Civita connection may be viewed as the Christoffel Symbol $\Gamma$ that defines the directional derivative in $M$.
\begin{equation}\label{eq.direc_direv}
D_{\partial x_{i}} \partial x_{j} = \sum_{k} \Gamma_{i,j}^{k} \partial x_{k}
\end{equation}
where $x_{i}$ is the coordinate function of an arbitrary point $m$ in $M$. The tangents of $M$ can be written as linear combination of $\{\partial x_{i}\}_{i=1,2,\cdots,dim(M)}$. For example at $X_{0}$, $V_{i}(X_{0})=\sum_{j} a_{j}^{i} \partial x_{j}$ and $dX_{0} = \sum_{j} b_{j} \partial x_{j}$.
Thus according to Equation $(\ref{eq.direc_direv})$, we can analytically calculate the derivative of $V_{i}(X_{t})$ along $X_{t}$ in direction $dX_{t}$, which is denoted as $D_{\frac{\partial X_{t}}{\partial
t}}(V_{i}(X_{t}))$. $V(X(t))$ is parallel if $\forall i$,
\begin{equation}
D_{\frac{\partial X_{t}}{\partial
t}}(V_{i}(X_{t})) = 0.
\end{equation}

However in practical problems, it is usually impossible to chart the manifold. In other words, the coordinate functions $x$ of $M$ are usually unknown. In this paper, the shape manifold \cite{KS04} we introduced in Section \ref{sec.shape_manifold} have some nice properties that allows us to calculate $D_{\frac{\partial X_{t}}{\partial
t}}(V_{i}(X_{t}))$ in terms of a projection of Euclidean calculus in the ambient space onto the tangent space of the manifold $M$.

\begin{equation}\label{eq.proj_deri}
D_{\frac{\partial X_{t}}{\partial
t}}(V_{i}(X_{t})) = \it{Proj}_{T_{X_{t}}M} \left( \frac{\partial \bar{V}_{i}(X_{t})}{\partial t} \right),
\end{equation}
where $\bar{V}_{i}$ is the vector representing $V_{i}$ in the ambient space of $M$. Since the ambient space is an affine space based on $\mathbb{L}^{2}$, thus $\bar{V}_{i}$ is calculable as a real vector in $\mathbb{L}^{2}$. The $\it{Proj}_{T_{X_{t}}M}$ is a mapping of vectors in $\mathbb{L}^{2}$ onto the tangent space of $M$.

\section{Dimension Reduction}\label{sec.dim_red}
The dynamics of some common human activities such as walking and
running, are relatively simpler in comparison to a complex activity
like dancing. Thus it is nature to speculate that the simple
activity could be represented in a lower dimensional subspace on the
shape manifold. To learn such sub-manifold, a dimension reduction is
proposed for a curve on a shape manifold. As described in
the previous section, first the horizontal vector field $V(X_{t})$
is defined by a Levi-Civita connection. Then Principle Component
Analysis (PCA) is carried out to compute the optimal vector field
$\tilde{V}(X_{t})= \{x_{t},e_{1},e_{2},\cdots,e_{L}\}$ such that all
the parallel transported tangents along $X_{t}$ could be represented
in a lower dimensional space.

The shape sequence on a manifold $M$ can be represented by
a differential equation as,
\begin{equation}\label{eq.mov_frame}
X_{t} = X_{0} + \int_{0}^{t} \sum_{i=1}^{dim(M)}
V_{i}(X_{t})dZ_{i}(t),
\end{equation}
where, $V(X_{t})$ is selected to be the horizontal vector field on a manifold and
$Z_{t}$ is the curve development in a Euclidean space. Such a
formulation have the property according to \cite{KN96,EH59} that the
resulting $Z_{t}$ is 1-1 corresponded to $X_{t}$ with the initial
condition $V(X_{0}),X_{0}$.

The idea of the proposed dimension reduction is to reduce the dimension for a manifold valued curve $X_{t}$ by learning a sequence of subspaces $\tilde{V}(X_{t})$ on the
shape manifold $M$, such that the resulting $dZ_{t}$ have a distribution that concentrate in a lower dimensional subspace.

A Levi-Civita connection is adapted on the shape manifold according to the $L^{2}$ metric induce from the ambient space. According to the definition of
covariant derivative under Levi-Civita connection as in Section \ref{sec.MFG}, the horizontal vector field $V(X_{t})$ is the solution of the following differential equation.
\begin{equation}\label{eq.pde}
\forall i, D_{\frac{\partial X_{t}}{\partial
t}}(V_{i}(X_{t}))=0,
\end{equation}
with initial condition $(X_{0},V(X_{0}))$.

The subspace moving frames $\tilde{V}(X_{t})$ for dimension reduction is constructed by select the optimal initial condition $V(X_{0}) = \tilde{V}(X_{0})$ . First parallel transport all the tangents $\frac{\partial X_{t}}{t}$ to the tangent space at $X_{0}$. With the moving frame formulation in Equation $(\ref{eq.mov_frame})$, the parallel transportation results is a set $\tau$
\begin{equation}
\tau = \{\sum_{i=1}^{dim(M)}
V_{i}(X_{t})dZ_{i}(t)\}_{t=1,2,\cdots,N}
\end{equation}
According to \cite{KN96}, the set $\tau$ is invariant to the choice of $V(X_{0})$. Thus in the first step an arbitrary initial condition is assigned.

Then Let the $\{e_{1},e_{2},e_{3},\cdots,e_{L}\}$ be the eigenvectors that corresponding to the $L$ largest eigenvalues of vectors in $\tau$. The optimal initial condition is constructed as
\begin{equation}
\tilde{V}_{i}(X_{0})= e_{i}.
\end{equation}
The solution $\tilde{V}(X_{t})$ of Equation $(\ref{eq.pde})$ with initial condition $(X_{0},\tilde{V}(X_{0}))$ is the moving frames we used to represent the $X_{t}$ by the corresponding $\tilde{Z}_{t}$ in a lower dimensional space $R^{L}$

In the following, we provide the solution to the differential equation $(\ref{eq.pde})$.
According to Equation $(\ref{eq.proj_deri})$ and normal space expression of the shape manifold $M$
in Section \ref{sec.shape_manifold}, the above equation can be solved as the following. Let $\{B_{1}(t),B_{2}(t),B{3}(t)\}$ be the orthogonalization  of $\{1,cos(X_{t}),sin(X_{t})\}$, which is the basis of normal space of the shape manifold $M$.  $\forall i$,
\begin{equation} \label{eq.cond1}
\frac{\partial V_{i}(X_{t})}{\partial t}  = a_{1}B_{1}(t_{0}) + a_{2}B_{2}(t_{0}) + a_{3}B_{3}(t_{0}).
\end{equation}
In the numerical calculation of the Euclidean derivatives $\left( \frac{ V_{i}(X_{t})}{\partial t} \right)$ are
implemented as follows. For a small enough $h>0$,
\begin{equation}\label{eq.direv_app}
\left( \frac{\partial V_{i}(X_{t})}{\partial t} \right) = \frac{ V_{i}(X_{t+h})- V_{i}(X_{t})}{h}.
\end{equation}
Substitute the Equation $(\ref{eq.direv_app})$ into the covariant derivative Equation $(\ref{eq.cond1})$, $V_{i}(X_{t+h})$ could be written as,
\begin{equation}\label{eq.par1}
V_{i}(X_{t+h}) = V_{i}(X_{t}) + a_{1}B_{1}(t_{0}) + a_{2}B_{2}(t_{0}) + a_{3}B_{3}(t_{0}).
\end{equation}
Since $V_{i}(X_{t+h}) \in T_{X_{t+h}}M$,  the parameter
$a_{i}$ should satisfy that,
\begin{equation*}
<V_{i}(X_{t}) + a_{1}B_{1}(t_{0}) + a_{2}B_{2}(t_{0}) + a_{3}B_{3}(t_{0}),
B_{1}(t_{0}+h) )> = 0,
\end{equation*}
\begin{equation*}
<V_{i}(X_{t}) + a_{1}B_{1}(t_{0}) + a_{2}B_{2}(t_{0}) + a_{3}B_{3}(t_{0}),
B_{2}(t_{0}+h)> = 0,
\end{equation*}
and
\begin{equation*}
<V_{i}(X_{t}) + a_{1}B_{1}(t_{0}) + a_{2}B_{2}(t_{0}) + a_{3}B_{3}(t_{0}), B_{3}(t_{0}+h)> = 0.
\end{equation*}
From the above three equations, we have,
\begin{equation}\label{eq.par2}
\left( \begin{array}{c}
   a_{1} \\
   a_{2} \\
   a_{3}
 \end{array}\right)
  = -A^{-1}V,
\end{equation}
where $A=$
{\tiny
\begin{equation*}\label{eq.par3}
\left(
      \begin{array}{ccc}
        B_{1}(X_{t}) B_{1}(X_{t+h}) & B_{2}(X_{t}) B_{1}(X_{t+h}) & B_{3}(X_{t})B_{1}(X_{t+h}) \\
        B_{1}(X_{t}) B_{2}(X_{t+h}) & B_{2}(X_{t}) B_{2}(X_{t+h}) & B_{3}(X_{t})B_{2}(X_{t+h}) \\
        B_{1}(X_{t}) B_{3}(X_{t+h}) & B_{2}(X_{t}) B_{3}(X_{t+h}) & B_{3}(X_{t})B_{3}(X_{t+h}) \\
      \end{array}
    \right).
\end{equation*} }
So the parameter $a_{i}$ could be solved in a linear
fashion.
Thus the shape sequence can be represented as follows,
\begin{equation}
X_{t} = X_{0} + \int_{0}^{t} \sum_{i=1}^{L}
\tilde{V}_{i}(X_{t})d\tilde{Z}_{i}(t),
\end{equation}
where $d\tilde{Z}_{i}(s)=<\frac{\partial X_{t}}{\partial t}, \tilde{V}_{i}(X_{t})> $ is a curve development in
$R^{L}$.

\begin{figure*}[htb]
\begin{minipage}[b]{1.0\linewidth}
  \centering
  \centerline{\epsfig{figure=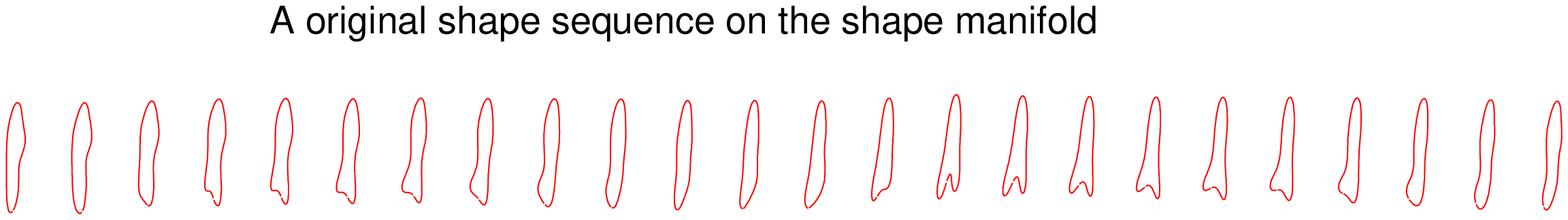,width=1\linewidth,height=0.2\linewidth}}
\end{minipage}
\caption{The original shape sequence for activity: Running}
\label{fig.ssq55}
\end{figure*}

\begin{figure*}[htb]
\begin{minipage}[b]{1.0\linewidth}
  \centering
  \centerline{\epsfig{figure=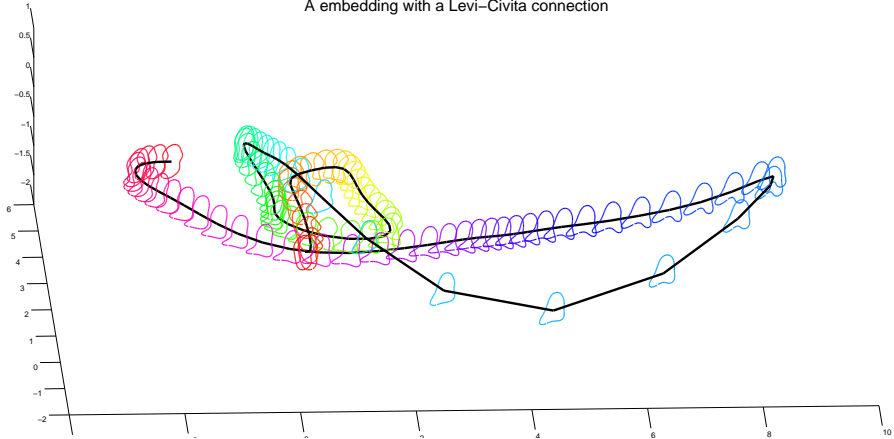,scale=0.3}}
\end{minipage}
\caption{The embedding of a shape sequence in Figure \ref{fig.ssq55}
into $R^{3}$} \label{fig.embedding55}
\end{figure*}

Figure \ref{fig.embedding55} illustrates the $R^{3}$ embedding of
the shape sequence in Figure \ref{fig.ssq55}. The curvature of the
embedded curve in $R^{3}$ in Figure \ref{fig.embedding55} represent
well the change of original shape sequence along time.

\begin{figure*}[htb]
\begin{minipage}[b]{1.0\linewidth}
  \centering
  \centerline{\epsfig{figure=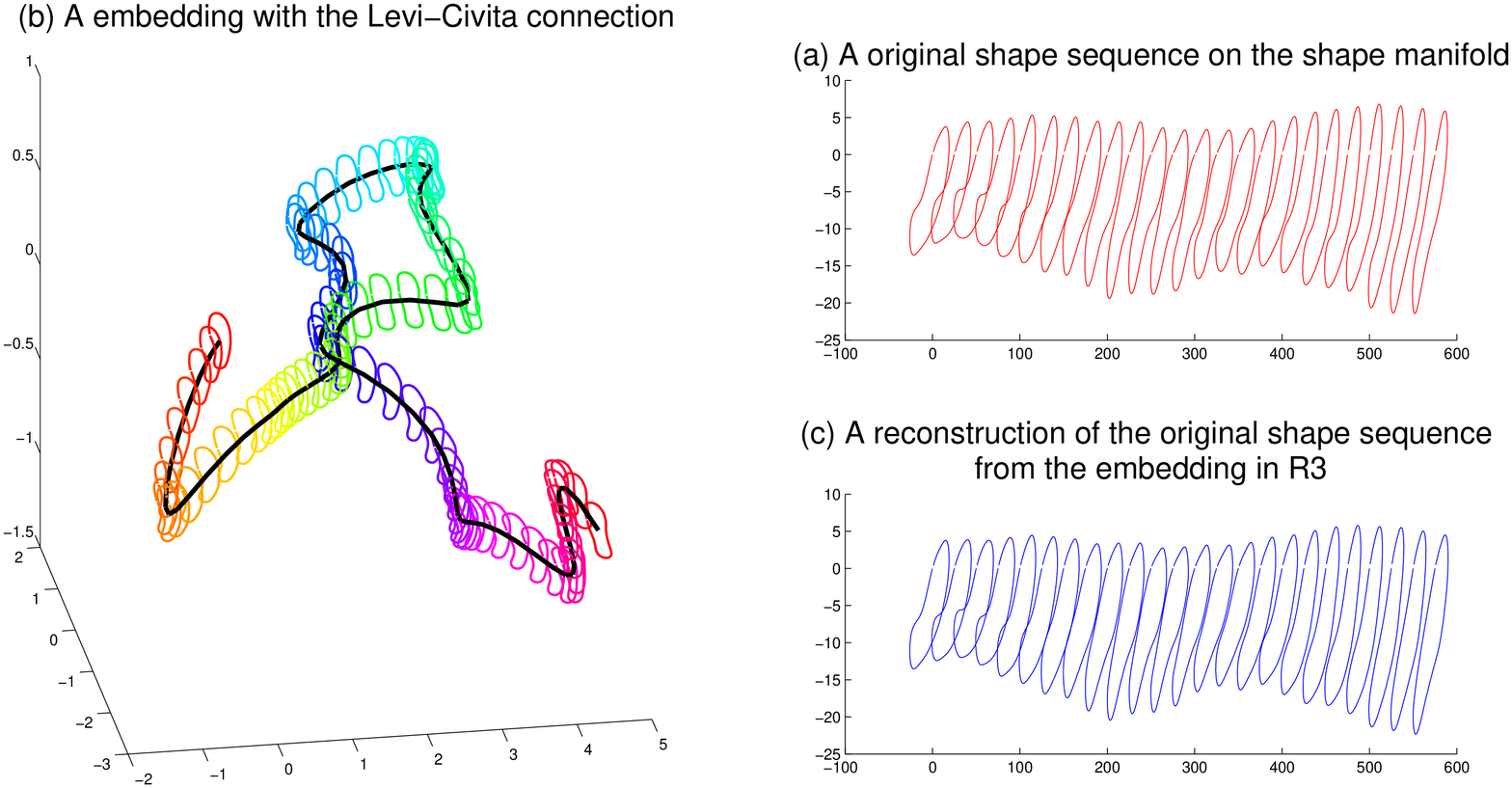, width=20cm,height = 5cm}}
\end{minipage}
\caption{(a) The original shape sequence; (b) The embedding into
$R^{3}$; (c) The reconstruction of the shape sequence from the
embedding curve in $R^{3}$.} \label{fig.embedding35}
\end{figure*}

\begin{figure*}[htb]
\begin{minipage}[b]{1.0\linewidth}
  \centering
  \centerline{\epsfig{figure=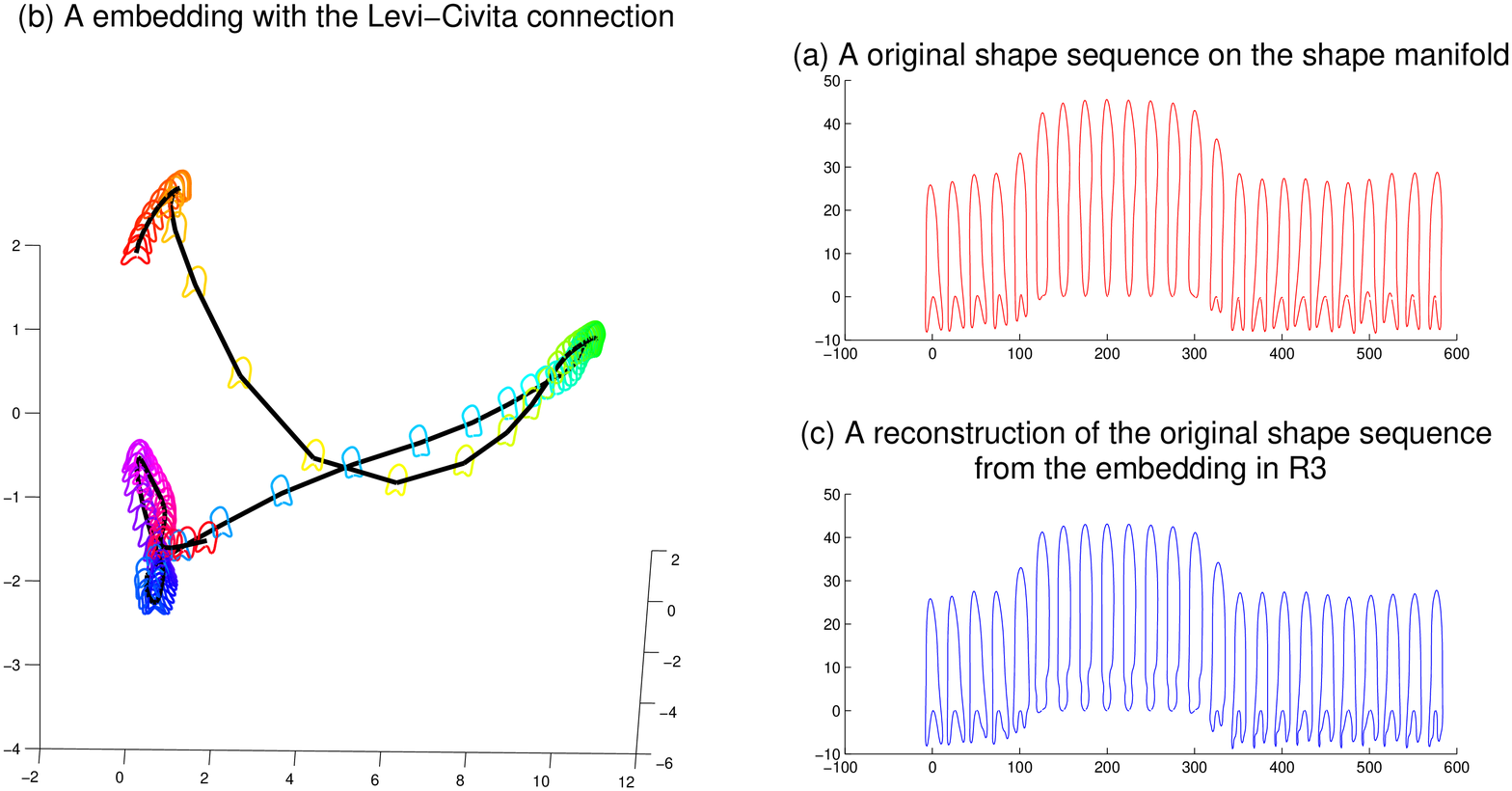, width=20cm,height = 5cm}}
\end{minipage}
\caption{(a) A original shape sequence; (b) The embedding into
$R^{3}$; (c) the reconstruction of the shape sequence from its
corresponding embedding curve in $R^{3}$.} \label{fig.embedding68}
\end{figure*}

\begin{figure*}[htb]
\begin{minipage}[b]{1.0\linewidth}
  \centering
  \centerline{\epsfig{figure=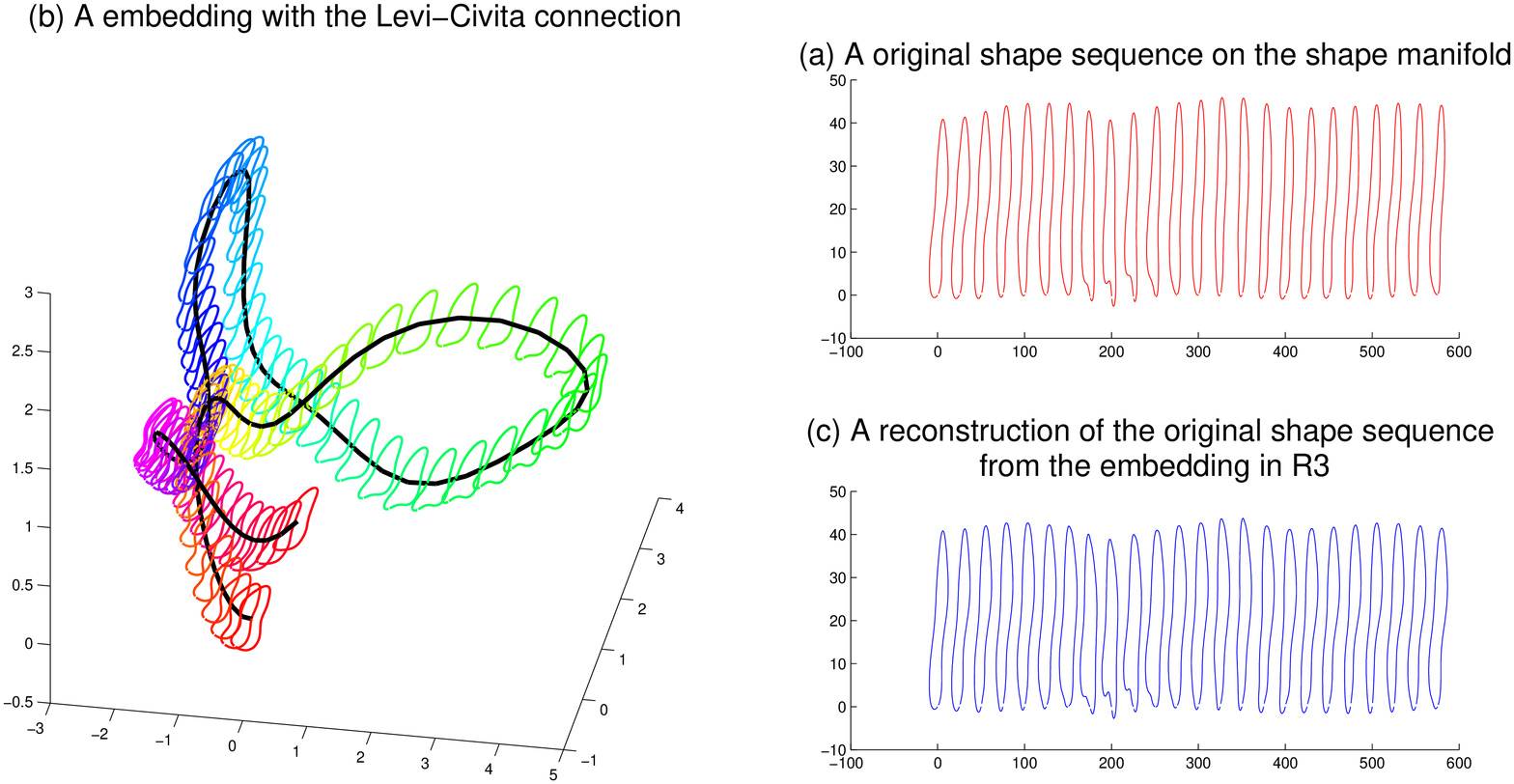, width=20cm,height = 5cm}}
\end{minipage}
\caption{(a) A original shape sequence; (b) The embedding into
$R^{3}$; (c) The reconstruction of the shape sequence from its
corresponding embedding curve in $R^{3}$.} \label{fig.embedding78}
\end{figure*}

\begin{figure*}[htb]
\begin{minipage}[b]{1.0\linewidth}
  \centering
  \centerline{\epsfig{figure=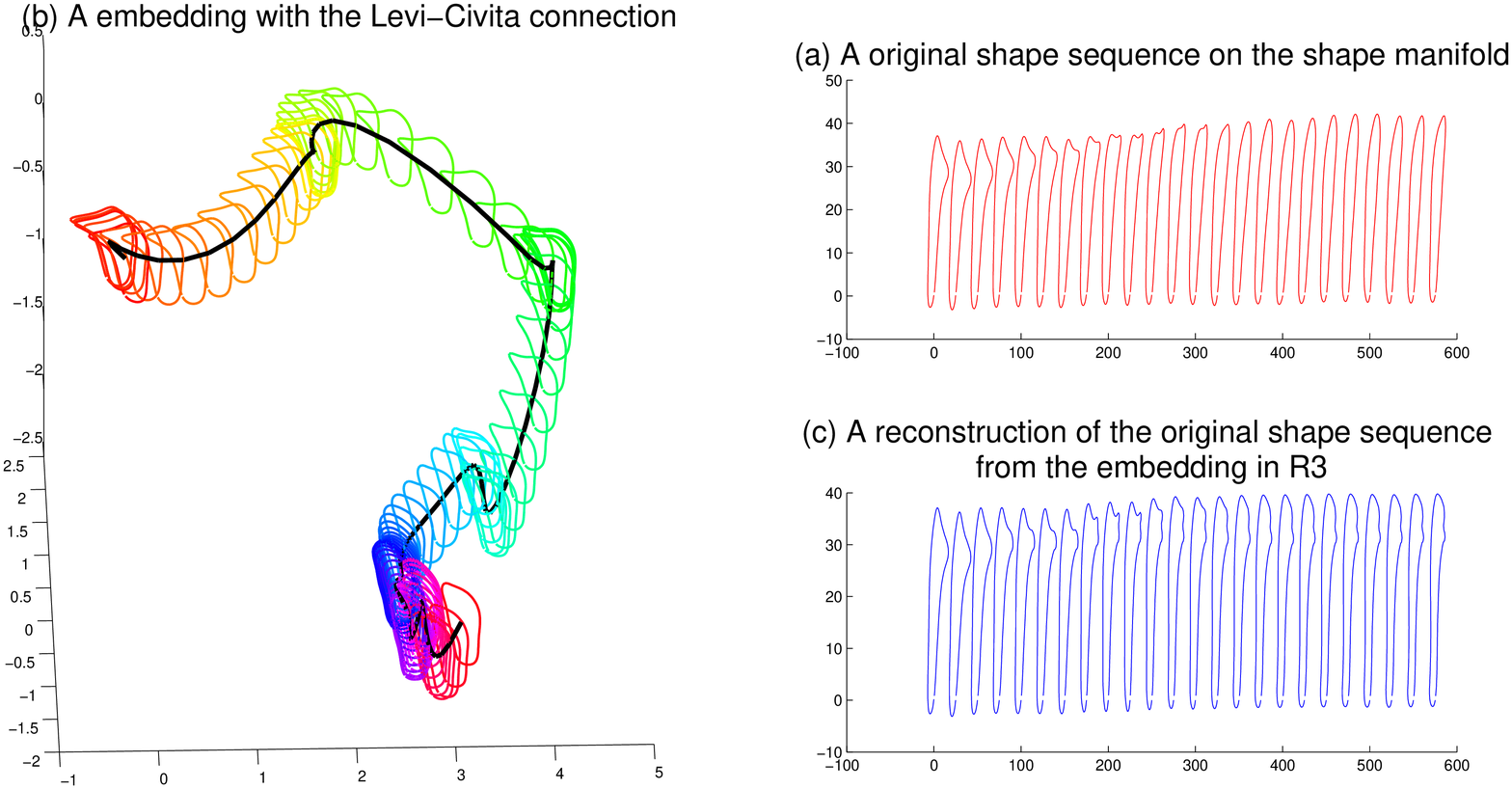, width=20cm,height = 5cm}}
\end{minipage}
\caption{(a) A original shape sequence; (b) The embedding into
$R^{3}$; (c) The reconstruction of the shape sequence from its
corresponding embedding curve in $R^{3}$.} \label{fig.embedding98}
\end{figure*}

In the following, the additional different activities from database \cite{BI05} is considered,
such as Jumping, Siding, Skipping and Waving. The dimension
reduction result in Figures
\ref{fig.embedding35},\ref{fig.embedding78},\ref{fig.embedding68},\ref{fig.embedding98},
also well visualized the evolution on the shape manifold $M$.

\section{Reconstruction from the Lower Dimensional Representation}\label{sec.dim_rec}
As introduced in the previous sections, one of the claimed advantages of the
proposed dimension reduction is that it is possible to generate the
original shape sequence from the dimension reduction results. Such
an invertible property makes our lower dimensional shape process
embedding a compelling way to learn the generative model of the
shape sequence.

Let $\tilde{Z}(t)$ be the embedded shape process in $R^{L}$. A
approximation of the original sequence on the shape manifold $M$ may
be generated by the following differential equation:

\begin{equation}\label{eq.gen1}
X_{t} = X_{0} + \int_{0}^{t} \sum_{i=1}^{L}
\tilde{V}_{i}(X_{0})d\tilde{Z}_{i}(t)
\end{equation}

where the horizontal vector field $\tilde{V}_{i}(X_{t}))$ is
generated uniquely by the differential equation \ref{eq.pde} with the initial frame
$X_{0},\tilde{V}_{i}(X_{0})$.

Numerically Equation $(\ref{eq.gen1})$ is implemented as a
difference function,
\begin{equation}\label{eq.gen2}
X_{k} = f( X_{k-1}) + \sum_{i=1}^{L}
\tilde{V}_{i}(X_{0})d\tilde{Z}_{i}(k-1))
\end{equation}

where $f$ is the iterative projection from the ambient space to the
shape manifold as in \cite{KS04}, which is also briefly introduced
in Section \ref{sec.shape_manifold}.

\begin{figure*}[htb]
\begin{minipage}[b]{1.0\linewidth}
  \centering
  \centerline{\epsfig{figure=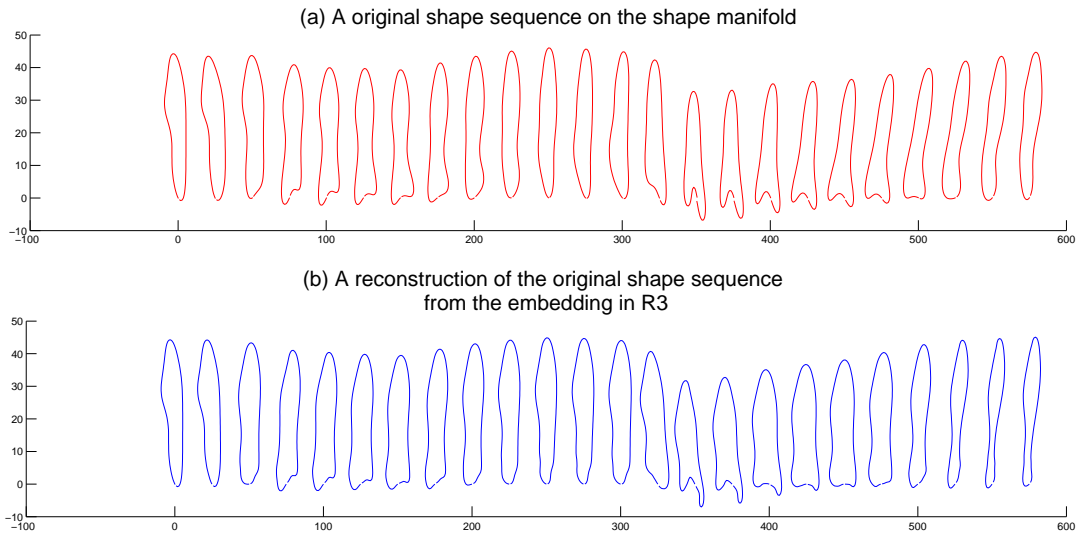,scale=0.4}}
\end{minipage}
\caption{The reconstruction result for original sequence in Figure
\ref{fig.ssq55} with embedding result in Figure
\ref{fig.embedding55}} \label{fig.rec55}
\end{figure*}

Figure \ref{fig.rec55} illustrates the reconstruction result of the
shape sequence of activity: running. The reconstruction is from the
embedding curve $\tilde{Z}(t) \in R^{3}$ and the initial condition
$X_{0}$, $\tilde{V}_{i}(X_{0})$ to the shape manifold $M$.

The Figure \ref{fig.rec55} demonstrates that the reconstruction
approximates well the original shape sequence. More reconstruction results of the shape sequence of other human activities are shown
in Figures \ref{fig.embedding35},\ref{fig.embedding78},\ref{fig.embedding68},\ref{fig.embedding98}.

\section{Conclusion}
Dimension reduction of points on a high dimensional manifold is well
studied in the past decades. Only a few works, however, address the
problem of dimension reduction of curves on a manifold. In this paper a novel dimension reduction technology
is proposed for shape dynamics with the following advantages that,
as far as we know, no previous work has ever achieved:
\begin{enumerate}
  \item The subspace is learned nonlinearly according to the geometry of the underlying manifold.
  \item The computation is linear in the size of data.
  \item There exists an efficient reconstruction from the lower dimensional
  representation to the original curve on a shape manifold.
  \item The proposed dimension
  reduction technique provides both analytical and practical
  foundations for generative modeling of shape dynamics.
\end{enumerate}
This work also first apply the differential geometry tools such as
moving frame representation and Levi-Civita connection to dimension reduction of curves on a manifold. It
provides a new perspective on the dimension reduction with a moving
frame formulation to naturally characterize the nonlinear features of
a manifold valued dynamics in a linear subspace.

\bibliographystyle{unsrt}

\end{document}